\documentclass[conference]{IEEEtran}

\usepackage{times}
\usepackage{epsfig}
\usepackage{graphicx}
\usepackage{amsmath}
\usepackage{amssymb}
\usepackage{subfigure}
\usepackage{multirow}
\usepackage{caption}
\usepackage{url}

\begin{document}
%
\title{Automated Inference on Sociopsychological Impressions of Attractive Female Faces}

\author{\IEEEauthorblockN{Xiaolin Wu}
\IEEEauthorblockA{McMaster University\\
Shanghai Jiao Tong University\\
Email: xwu510@gmail.com}
\and
\IEEEauthorblockN{Xi Zhang}
\IEEEauthorblockA{Shanghai Jiao Tong University\\
Email: zhangxi\_19930818@sjtu.edu.cn}
\and
\IEEEauthorblockN{Chang Liu}
\IEEEauthorblockA{Shanghai Jiao Tong University\\
Email: liu123@sjtu.edu.cn}}

\maketitle

\begin{abstract}
This article is a sequel to our earlier work \cite{wuzhang2016}.  The main objective of our research is to explore the  potential of supervised machine learning in face-induced social computing and cognition, riding on the momentum of much heralded successes of face processing, analysis and recognition on the tasks of biometric-based identification.  We present a case study of automated statistical inference on sociopsychological perceptions of female faces controlled for race, attractiveness, age and nationality.
Our empirical evidences point to the possibility of training machine learning algorithms, using example face images characterized by internet users, to predict perceptions of personality traits and demeanors.
\end{abstract}

\IEEEpeerreviewmaketitle

\section{Introduction}
One of the big successful stories of artificial intelligence is automatic face recognition.  Owning to years of extensive research on machine learning, face image processing and analysis \cite{alexnet, park1991, sobottka1998, thomaz2010, bartlett1998}, we are about to cross the technology hurdles and embrace a wide use of algorithms and systems that can determine a person's gender, race, age, identity and even emotion, based on the face image(s) of the person in question.  The next tantalizing and challenging question is whether supervised machine learning can draw statistical inferences on sociopsychological perception, personality traits, and behavioral propensity of individuals from face images.  This possibility cannot be denied without thorough examinations simply because of societal taboos and political beliefs. People in all cultures do acknowledge the effects of and are influenced by the ``first impression".   Here a Turing test of a sort presents itself: can artificial intelligence duplicate the face-induced perceptions pertaining to social aspect of humans?

Recently we launched a new line of inquiry to examine and explore the potential of machine learning in statistically inferring social perceptions of a face image, such as personality traits and demeanors, to
non-acquaintances of a particular social group delimited by race, gender, age, political belief, social value, etc.\  At the onset of this research, we made it very clear, as stated in \cite{wuzhang2016}, that our interests are in finding machine learnable correlations, if any, between facial appearances and social attributes of the observed and the observer, {\em not} in the interpretations of our findings in the realm of social sciences.
It will be remarkable and even somewhat surprising if machine learning is found to be capable of acquiring human-like first impressions of total strangers, because even humans have difficulties in rationalizing their first impression of others' social attributes and in precisely characterizing facial features that induce their perceptions.


We started our investigation on what we thought a relatively easy case: automated face-induced statistical inference on the propensity of law breaking \cite{wuzhang2016}.  Law-breakers and law-abiding people constitute the two populations that are arguably furthest apart in social attributes and behavioral predisposition.  In a multitude of social dimensions, such as self-control, trustworthiness, dominance and innocence, law breakers, particularly the violent type, tend to populate the perimeters of the multidimensional distribution.  If correlation exists between facial appearance and innate personal traits, then it should not be a surprise that law-breakers and law-abiding people are statistically separable by automatic face classifiers built by supervised machine learning, as turned out in our earlier study.  In this work, a sequel to \cite{wuzhang2016},  we take face-induced automatic inference of social attributes to what appears to be a more elusive task: classifying different styles of female attractiveness, which are either approved or not approved by certain disjoint subsets of observers.

Female attractiveness has been studied by researchers in psychology \cite{van1962, cunningham1986, langlois1990, langlois1994, perrett1994, manning1997, moller1995, cunningham1995, rhodes2000, fink2006, lie2008, little2011}.  But in all previous experiments human judges' opinions were solicited.
The prevailing view is that the perception of female beauty is primarily derived from cues of health and reproductive potential.  Most papers on facial attractiveness of female focus on its physical and physiological characteristics \cite{van1962, thornhill1993, roberts2005, lie2008, langlois1990, langlois1994}.

However, female attractiveness is also a social perception, as reflected by the cliche phrase ``beauty is in the eye of the beholder''.  A non-acquaintance female face can be judged unanimously as being physically beautiful, and yet different observers may associate this face with approval or disapproval connotations, which may vary in different cultural and/or historical contexts, using labels (or stereotypes) like \emph{pure}, \emph{sweet}, \emph{endearing}, \emph{innocent}, \emph{cute}, ... on one hand, or \emph{indifferent}, \emph{pretentious}, \emph{pompous}, \emph{arrogant}, \emph{frivolous}, \emph{coquettish}, ... on the other.
As these labels are loosely binary quantization of social attributes of trustworthiness, dominance, innocence, and extroversion, here is another case for the old, cross-culture belief that facial appearances are symptoms of innate traits and behavioral propensity.

Built upon our progress in automated statistical inference on the face-induced perception of the propensity of law breaking, albeit being controversial, we take the same approach of supervised machine learning, and examine whether a trained data-driven classifier can distinguish different styles of facial attractiveness of females as discussed above.  This classification problem, to humans, is not any easier than separating law breakers from law-abiding persons. The perception of facial attractiveness is more complex than the already knotty matter of subjective taste, for it is also a compound proxy for personalities and social values of the observed and the observer.

The remainder of this paper is structured as follows.  In Section II, we detail a semi-automatic process of collecting and preparing sample data, which is necessary for setting up the experiments, with control variables of race, attractiveness, age and nationality, to assess the potential of the supervised machine learning in statistical inference on sociopsychological perceptions of female faces.
In Section III, we teach a CNN classifier, using almost 4000 attractive face images of young Chinese females, each of which is attached to one or more sociopsychological label(s), to predict Chinese young men's perception of a young woman in terms of her personality traits and demeanors.
The experimental results turn out to be quite encouraging, and we offer and invite discussions and interpretations.  Section IV concludes the paper.

\section{Data Preparation}
In the interests of the availability of a large sample set for training and testing, and also of having our experiments controlled for attractiveness, race, gender and age, our research subjects are restricted to young Chinese females that are considered by mainstream Chinese to have attractive faces.  Controlling the variable of attractiveness is crucial for the validity of our study because it strongly influences a non-acquaintance's first impression of the female.

We run the Baidu image search engine with key words beautiful/pretty/attractive girls/young women, to select the sample images.  The Baidu image search engine relies mainly on captions, labels and blogs associated with the images to produce the query results.  The selected face images are divided into two subsets, denoted by $S_1$ and $S_0$, corresponding to the following two different types of demeanors and/or attitudes.

The set $S_1$ contains the images returned by the Baidu image search engine when it is requested to find attractive girls but with qualifiers:
\begin{itemize}
\item \emph{sweet, endearing, elegance, tender, caring, cute}.
\end{itemize}

In contrast, the set $S_0$ contains the images returned by the Baidu image search engine when it is requested to find attractive girls but with qualifiers:
\begin{itemize}
\item \emph{pretentious, pompous, indifferent, coquettish}.
\end{itemize}

Note that the results of the qualifier-based query with the Baidu image search engine cannot be directly used for our purpose.  The key words attached to the images cannot be taken at face value.  For instance,
an image associated with comment ``a girl beautiful but not sweet" can be a match for a query with combined key words ``beautiful sweet girl", which erroneously places an instance in subset $S_0$ into subset $S_1$.  To correct this problem, we use a simple text pattern matching program to detect such cases where the adjective is after a negation prefix.  Finally, all images passing the software selection and screening process are examined manually by male Chinese graduate students to detect selection errors due to more difficult semantics of the image-associated text such as sarcastic remarks.  Even though we have made every effort possible within our resources to ensure the data quality, there may be a very small probability of label noises.

Also, there is a caveat for anyone who wants to interpret our results down the road.  Not all labels of social perceptions attached to the sample images are given by non-acquaintances.  A small percentage of labels are apparently the results of observing the person in question for a period of time, hence they may carry some more information than others.

%

\begin{figure}
\centering
\includegraphics[width=0.18\columnwidth]{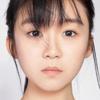}
\includegraphics[width=0.18\columnwidth]{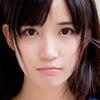}
\includegraphics[width=0.18\columnwidth]{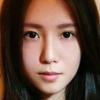}
\includegraphics[width=0.18\columnwidth]{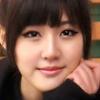}
\includegraphics[width=0.18\columnwidth]{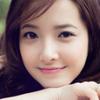}
\\[0.5ex]
\includegraphics[width=0.18\columnwidth]{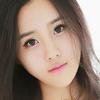}
\includegraphics[width=0.18\columnwidth]{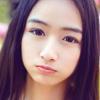}
\includegraphics[width=0.18\columnwidth]{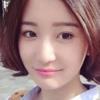}
\includegraphics[width=0.18\columnwidth]{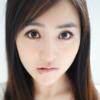}
\includegraphics[width=0.18\columnwidth]{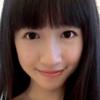}
\\[0.5ex]
\includegraphics[width=0.18\columnwidth]{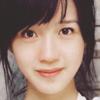}
\includegraphics[width=0.18\columnwidth]{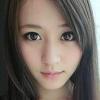}
\includegraphics[width=0.18\columnwidth]{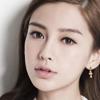}
\includegraphics[width=0.18\columnwidth]{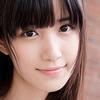}
\includegraphics[width=0.18\columnwidth]{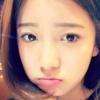}

\caption{The sample face images in subset $S_1$.}
\label{sample_p}
\end{figure}

\begin{figure}
\centering
\includegraphics[width=0.18\columnwidth]{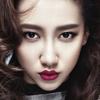}
\includegraphics[width=0.18\columnwidth]{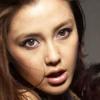}
\includegraphics[width=0.18\columnwidth]{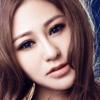}
\includegraphics[width=0.18\columnwidth]{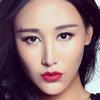}
\includegraphics[width=0.18\columnwidth]{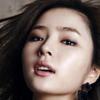}
\\[0.5ex]
\includegraphics[width=0.18\columnwidth]{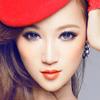}
\includegraphics[width=0.18\columnwidth]{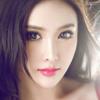}
\includegraphics[width=0.18\columnwidth]{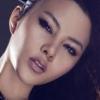}
\includegraphics[width=0.18\columnwidth]{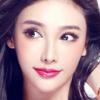}
\includegraphics[width=0.18\columnwidth]{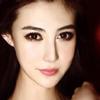}
\\[0.5ex]
\includegraphics[width=0.18\columnwidth]{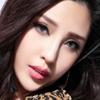}
\includegraphics[width=0.18\columnwidth]{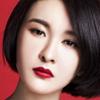}
\includegraphics[width=0.18\columnwidth]{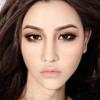}
\includegraphics[width=0.18\columnwidth]{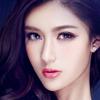}
\includegraphics[width=0.18\columnwidth]{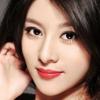}

\caption{The sample face images in subset $S_0$.}
\label{sample_n}
\end{figure}

Through the above semi-automatic data gathering and screening process, we collect 3954 photographs of attractive young Chinese females, among which 2000 belong to subset $S_1$ and 1954 belong to subset $S_0$.  The selected test photographs are processed by cloud service Face++ \cite{facepp}.  Face++ identifies and extracts the face(s) out of each photograph.  The resulting face images are scaled to $110 \times 110$ resolution to be ready for supervised machine learning.  Figures \ref{sample_p} and \ref{sample_n} list some sample face images in both subsets $S_1$ and $S_0$.


To our best knowledge, the internet users who attach the captions, labels and comments to the photographs of attractive Chinese females posted on line are predominantly young Chinese males.
As such, the two classes of female face images in $S_1$ and $S_0$ reflect esthetic preference and value judgments that prevail among young males in contemporary China.   Therefore, if it can classify between $S_1$ and $S_0$ based on face images with sufficiently high probability, then the supervised machine learning once more demonstrates its potential in drawing statistical inference on social attributes and behavioral propensity from face images.  Indeed, the significance of this study lies in collecting empirical evidences for the capability of machine learning, or lack of it, to perform sociopsychological cognition tasks, which appears to be at a degree higher and subtler than the determination of biometric attributes (e.g., gender, race, age and unique facial signatures for identification) \cite{geng2007, moghaddam2000, makinen2008, levi2015}.

\begin{table} \normalsize
\centering
\begin{tabular}{|c|c|c|c|}
\hline
Attributes & Dominance & Trustworthiness & innocence \\
\hline
\hline
\emph{Sweet} & 0 & 1 & 1 \\
\emph{Endearing} & 0 & 1 & 1 \\
\emph{Elegance} & 0.5 & 1 & 0.5 \\
\emph{Caring} & 0 & 1 & 0.5 \\
\emph{Cute} & 0 & 1 & 1\\
\emph{Tender} & 0 & 1 & 1\\
\hline
\hline
\emph{Pretentious} & 0.5 & 0 & 0 \\
\emph{Pompous} & 1 & 0.5 & 0 \\
\emph{Indifferent} & 1 & 0.5 & 0 \\
\emph{Coquettish} & 0.5 & 0 & 0 \\
\hline
\end{tabular}
\caption{Quantization of labels pertaining to $S_1$ and $S_0$ in the attribute space.}
\label{quantize}
\end{table}

In the resulting sets of sample images, each of the key words used in the image queries can be considered as vector quantization of three-dimensional social attributes (dominance, trustworthiness, innocence).  Specifically, we quantize each query key word in the attribute space of (dominance, trustworthiness, innocence) as in Table \ref{quantize}.

\begin{figure}
\centering
\includegraphics[width=\columnwidth]{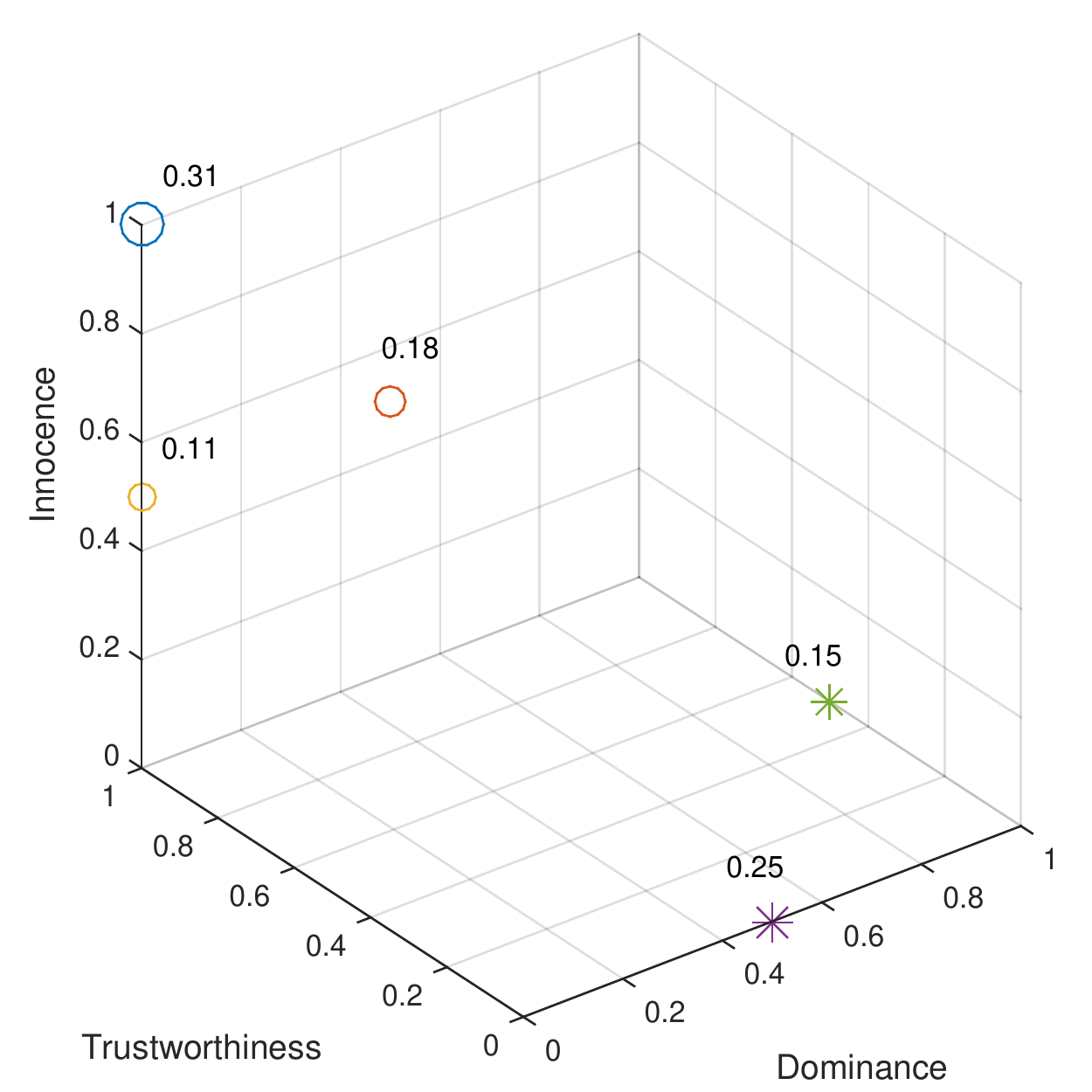}
\caption{Sample distributions of the subset $S_1$ and subset $S_0$ in the three-dimensional space of social perceptions.}
\label{distri}
\end{figure}

Figure \ref{distri} plots the sample distributions of subsets $S_1$ and $S_0$ in the above three-dimensional space of social perceptions.  The immediate observation is that the two subsets of attractive young Chinese females have strong negative correlation in the vector of social perceptions, or at the opposite ends in terms of perceived personalities, demeanors and attitudes.   Such polarized social perceptions are apparently caused by facial appearances because the face is the main object in all selected images.

\section{Experiments and Results}

In this section, we examine whether the supervised machine learning can acquire the personal taste and social value of the mainstream Chinese male internet users who label/caption the sample images of attractive Chinese females.  If yes, this adds a hint of evidence for the capability of machine learning for social cognition tasks.



Although the face-induced sociopsychological impressions of attractive Chinese young females left on mainstream Chinese young males appear to be quite consensual in the observers group, these young men have difficulty in pinpointing exactly what facial features and head poses contribute to their perceptions.
We show our sample images in both subsets $S_1$ and $S_0$ to 22 male Chinese graduate students, their perceptions of these face images agree with the labels (the mainstream stereotypes) given by the internet users.  But when being pressed to explain their judgments with specifics, they all give vague answers like ``I just feel this way".  In this case, the most appropriate tool of machine learning for automatic inference on sociopsychological impressions of faces is the Deep Convolutional Neural Network (CNN), which does not need to operate on explicit features like in SVM, KNN and other machine learning methods.

\begin{figure*}
\centering
\includegraphics[width=\textwidth]{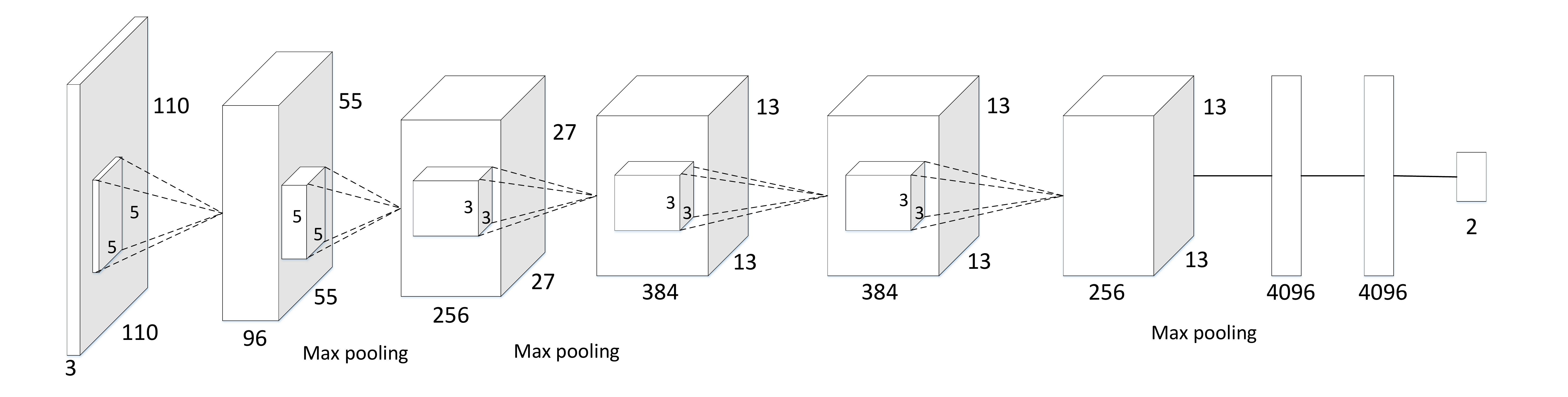}
\caption{The architecture of the AlexNet tailored to our problem.}
\label{alexnet}
\end{figure*}

We choose one of the well-known CNN architectures widely known as AlexNet \cite{alexnet}, and train it to identify which style of female attractiveness, represented by $S_1$ or $S_0$, a face image likely belongs to; namely, predicting the social perception of an attractive Chinese female, whether she will be perceived favorably by those young Chinese men who prefer a traditional type of personality traits, demeanor, and attitude.
Figure \ref{alexnet} illustrates the architecture of the AlexNet tailored to our problem.  The AlexNet contains eight layers with weights; the first five are convolutional and the remaining three are fully-connected.  We only borrow the architecture of the AlexNet, and train all the parameters of every layer in AlexNet for the binary classification task of discriminating between $S_1$ and $S_0$.

\begin{figure}
\centering
\includegraphics[width=\columnwidth]{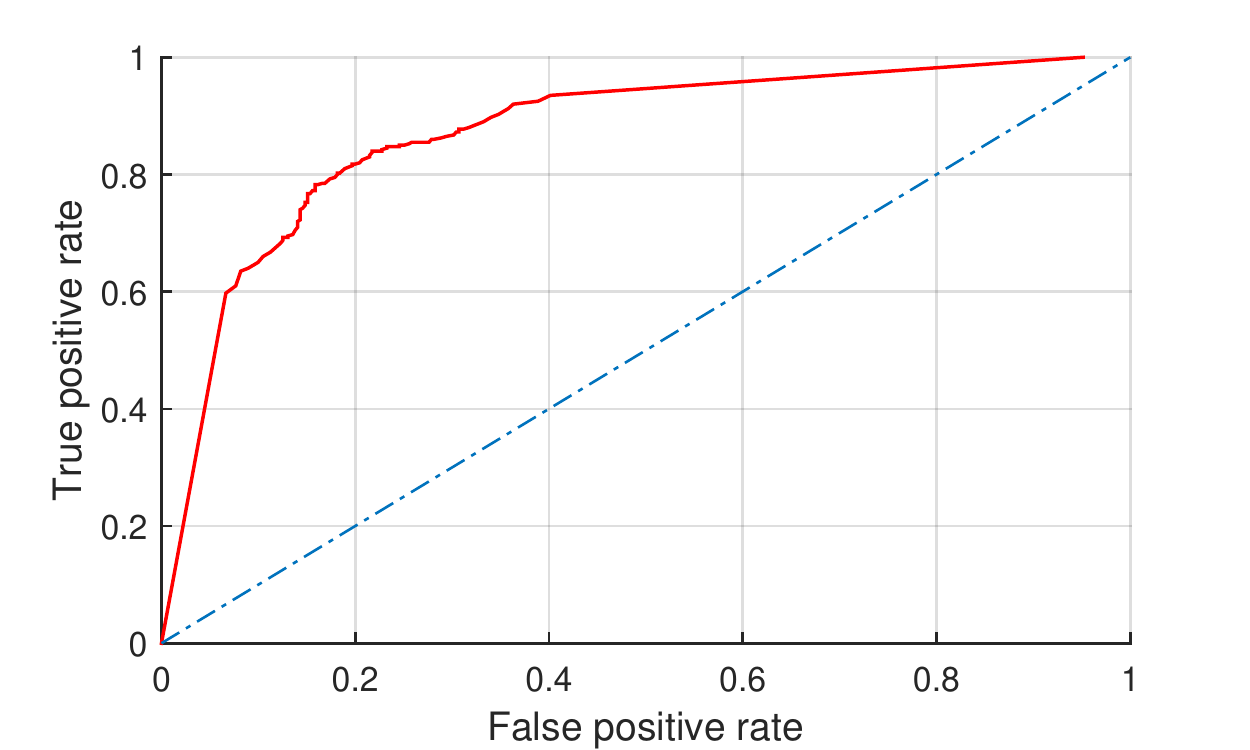}
\caption{The ROC curve of our neural network for statistical inference on styles of female attractiveness.}
\label{ROC}
\end{figure}

In our experiments, 80\% of the face images in our data set are used for training, 10\% of the face images for validation, and the remaining 10\% of face images for testing.  This training process is repeated ten times for different random seeds.
Table \ref{res} tabulates the accuracy, false positive rate and false negative rate of our neural network for statistical inference on styles of female attractiveness.  The ROC curve for our data-driven face classifier is also presented in Figure \ref{ROC}.

\begin{table} \normalsize
\begin{center}
\begin{tabular}{|c|c|c|}
\hline
Accuracy & False Positive Rate & False Negative Rate \\
\hline
\hline
80.23\% & 21.05\% & 19.65\% \\
\hline
\end{tabular}
\end{center}
\caption{The accuracy, false positive rate and false neagtive rate of our neural network for statistical inference on styles of female attractiveness.}
\label{res}
\end{table}

As indicated by Table \ref{res} and Figure \ref{ROC}, the CNN face classifier performs well in inference on sociopsychological perception of attractive Chinese young females.  This is quite remarkable given the fact that even human observers (male Chinese graduate students) have difficulty with rationalizing their sociopsychological perceptions of the tested female face images.  Unfortunately, the CNN face classifier, effective as it is, does not offer any explanations for its success either.

\begin{figure}
\centering
\includegraphics[width=\columnwidth]{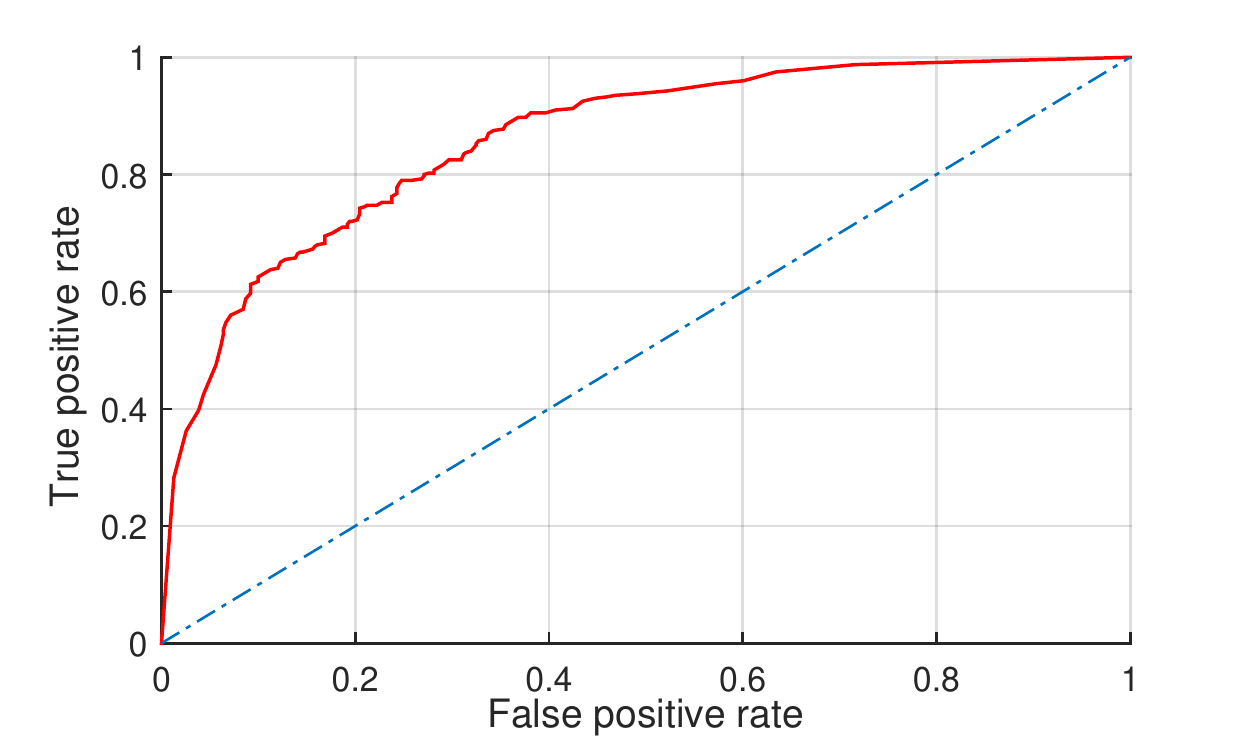}
\caption{The ROC curve of the proposed CNN classifier for grey scale face images.}
\label{ROC_grey}
\end{figure}

One comment that is frequently made by Chinese male graduate students to justify their approval of some female face images is ``look natural".   This leads us to suspicion that the CNN face classifier uses heavy facial makeups as an important cue in its decision.  To verify our hunch, we remove colors and use only grey scale face images to train and test the CNN classifier.  The deprivation of colors will reduce the effects of facial makeups because the use of colors is critical in female cosmetics.  Our experiments with grey scale face images turn out to be inconclusive.  The accuracy of the CNN face classifier decreases by only 6\% without color information.
Table \ref{res_grey} tabulates the accuracy, false positive rate and false negative rate of our neural network for grey scale face images.  The ROC curve for grey scale face images is also presented in Figure \ref{ROC_grey}.

\begin{table} \normalsize
\begin{center}
\begin{tabular}{|c|c|c|}
\hline
Accuracy & False Positive Rate & False Negative Rate \\
\hline
\hline
74.59\% & 24.81\% & 26.00\% \\
\hline
\end{tabular}
\end{center}
\caption{The accuracy, false positive rate and false negative rate of the proposed CNN classifier for grey scale face images.}
\label{res_grey}
\end{table}

\begin{figure}
\centering
\subfigure[Histogram of contrast.]{\label{contrast}
\includegraphics[width=\columnwidth]{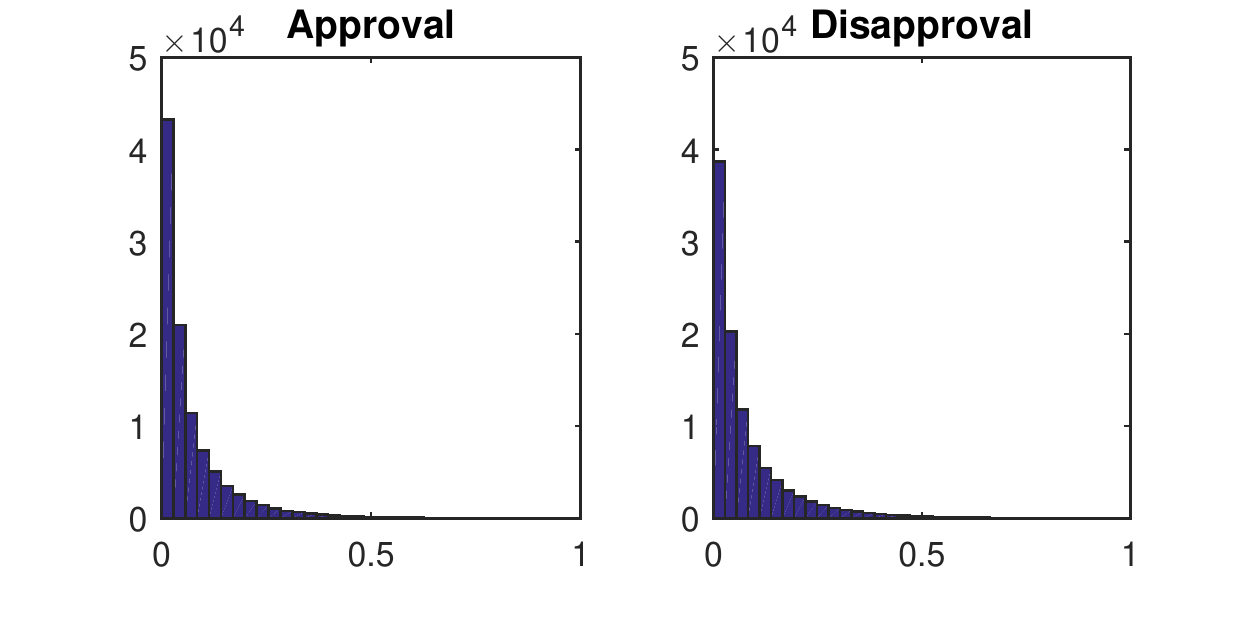}
}
\subfigure[Histogram of saturation.]{\label{saturation}
\includegraphics[width=\columnwidth]{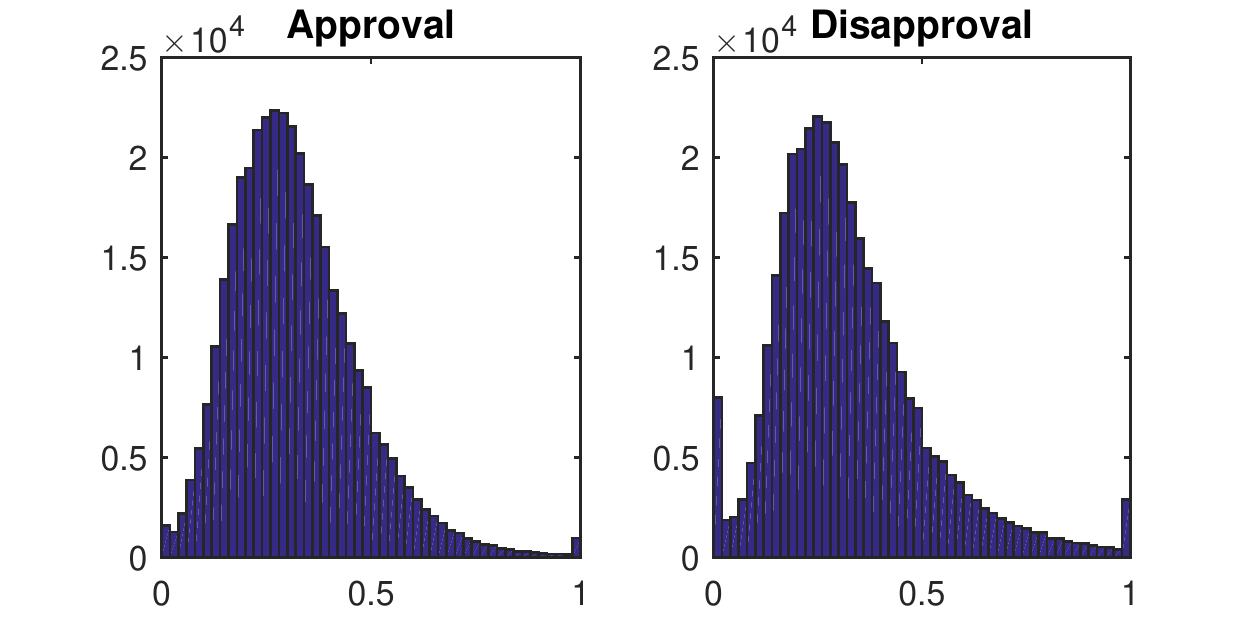}
}
\caption{Histograms of color contrast and saturation for subset $S_1$ and $S_0$.}
\label{hist}
\end{figure}

Other symptoms of heavy makeups include overuse of color saturated lipsticks, mascara, eyebrow pencils, etc.\  Therefore, we compare the distributions of color contrast and saturation for face images in $S_1$ and $S_0$.  Figure \ref{hist} presents the histograms of color contrast and saturation for $S_1$ and $S_0$ after normalizing measurements to range $[0,1]$.  The mean and standard deviation of color contrast and saturation for $S_1$ and $S_0$ are tabulated in Table \ref{mean}.  As expected, the color contrast of $S_1$ is on average 13.84\% smaller than that of $S_0$; the color saturation of $S_1$ is on average 4.85\% smaller than that of $S_0$;  Interestingly, the saturation histogram for $S_0$ has two distinctive spikes at the two extreme ends, indicating that women in subset $S_0$ tend to use thick black (saturation = 0) or/and saturated (vivid) colors. We also note that the variations of color contrast and saturation of face images in $S_0$ are appreciably greater than those of $S_1$.  This apparently reflects the traditional esthetics preference for and social value of naturalness in Chinese culture.  Such quite subtle cues in sociopsychological aspect seem to be picked up by our CNN method, otherwise it would be difficult to explain the good performance of the proposed face classifier.

\begin{table} \normalsize
\centering
\begin{tabular}{|c|c|c|c|c|}
\hline
 \multicolumn{2}{|c|}{Impressions} & $S_1$ & $S_0$ \\
\hline
\multirow{2}{*}{Mean} & Contrast & 0.0708 & 0.0806 \\
 & Saturation & 0.3195 & 0.3350 \\
\hline
\hline
\multirow{2}{*}{Standard deviation} & Contrast & 0.0910 & 0.1013 \\
 & Saturation & 0.1509 & 0.1784 \\
\hline
\end{tabular}
\caption{The mean and standard deviation of color contrast and saturation for subset $S_1$ and $S_0$.}
\label{mean}
\end{table}

Finally, to safe guard against possible risk of data overfitting by our neural network for statistical inference on sociopsychological perceptions of attractive female faces, we conduct the fault-finding experiment proposed in \cite{wuzhang2016}, in seeking for counterexamples.
We randomly label the faces in our sample set as positive and negative instances with equal probability, and redo the above experiments of classification after retraining the CNN with the artificially labeled samples. The experimental results show that the randomly generated positive and negative instances cannot be distinguished at all by the CNN method; the average classification accuracy, false positive rate and false negative rate are all about 50\%.  Based on these findings, we are confident that the good accuracy of the proposed CNN face classifier is not due to data overfitting; otherwise, given the same sample size, the trained classifier would also be able to statistically separate the randomly labeled positive instances from the negative ones, rather than guessing at random as found in the above experiment.

\section{Conclusion}

This work is a sequel to our earlier paper \cite{wuzhang2016}.  We drive the research on face processing, analysis and recognition beyond the tasks of biometric-based identification, and try to extend it in the direction of automatic statistical inferences on sociopsychological perceptions,
such as personality traits and behavioral propensity.  By the reported case study on face attractiveness of young Chinese females, we demonstrate once again the potential of supervised machine learning in face-induced social computing and cognition.

{
\bibliographystyle{ieee}
\bibliography{egbib}
}

\end{document}